\documentclass[10pt,twocolumn,letterpaper,table]{article}

\usepackage[pagenumbers]{arxiv} 

\usepackage{caption,subcaption}

\usepackage{array}
\usepackage{amsmath}
\usepackage{times}
\usepackage{epsfig}
\usepackage{graphicx}
\usepackage{amsmath}
\usepackage{amssymb}
\usepackage{physics}

\usepackage{epsfig}
\usepackage{epstopdf}
\usepackage{booktabs}
\usepackage{algpseudocode} 
\usepackage{multirow}
\usepackage{textcomp}
\usepackage{gensymb}
\usepackage{pifont}
\usepackage{bm}

\usepackage{tabularx}
\usepackage{adjustbox}
\usepackage{array}
\usepackage{multirow}
\usepackage{url}
\usepackage{array}
\usepackage{lipsum}
\usepackage{makecell} 
\usepackage{enumitem}
\usepackage{algorithm}

\usepackage[table]{xcolor}
\DeclareMathAlphabet\mathbfcal{OMS}{cmsy}{b}{n}

\newcolumntype{R}[2]{%
    >{\adjustbox{angle=#1,lap=\width-(#2)}\bgroup}%
    l%
    <{\egroup}%
}

\definecolor{baselinecolor}{gray}{.9}
\definecolor{deemph}{gray}{0.6}

\definecolor{arxivcolor}{rgb}{0.21,0.49,0.74}
\usepackage[pagebackref,breaklinks,colorlinks,citecolor=arxivcolor]{hyperref}

\title{MoEC: Mixture of Experts Implicit Neural Compression}
\author{
    Jianchen Zhao$^{1,}$\thanks{Equal Contribution, $^\dagger$Corresponding Author}
    \qquad \quad
    Cheng-Ching, Tseng$^{1,*}$\qquad \quad 
    Ming Lu$^{1}$\qquad \quad
    Ruichuan An$^{1}$\qquad \quad \\
    Xiaobao Wei$^{1}$\qquad \quad
    He Sun$^{2,3}$\qquad \quad
    Shanghang Zhang$^{1,\dagger}$
    \\
    $^{1}$National Key Laboratory for Multimedia Information Processing,\\ School of Computer Science, Peking University \\
    $^{2}$National Biomedical Imaging Center, Peking University, Beijing, China. \\
    $^{3}$College of Future Technology, Peking University, Beijing, China. \\
    {\tt\small shanghang@pku.edu.cn}\\
}

\begin{document}
\maketitle
\begin{abstract}
Emerging Implicit Neural Representation (INR) is a promising data compression technique, which represents the data using the parameters of a Deep Neural Network (DNN). Existing methods manually partition a complex scene into local regions and overfit the INRs into those regions. However, manually designing the partition scheme for a complex scene is very challenging and fails to jointly learn the partition and INRs. To solve the problem, we propose MoEC, a novel implicit neural compression method based on the theory of mixture of experts. Specifically, we use a gating network to automatically assign a specific INR to a 3D point in the scene. The gating network is trained jointly with the INRs of different local regions. Compared with block-wise and tree-structured partitions, our learnable partition can adaptively find the optimal partition in an end-to-end manner. We conduct detailed experiments on massive and diverse biomedical data to demonstrate the advantages of MoEC against existing approaches. In most of experiment settings, we have achieved state-of-the-art results. Especially in cases of extreme compression ratios, such as 6000x, we are able to uphold the PSNR of 48.16. The code will be released.
\end{abstract}    
\section{Introduction}
\label{sec:intro}

\begin{figure*}[t]
\centering
\includegraphics[width=1.\textwidth]{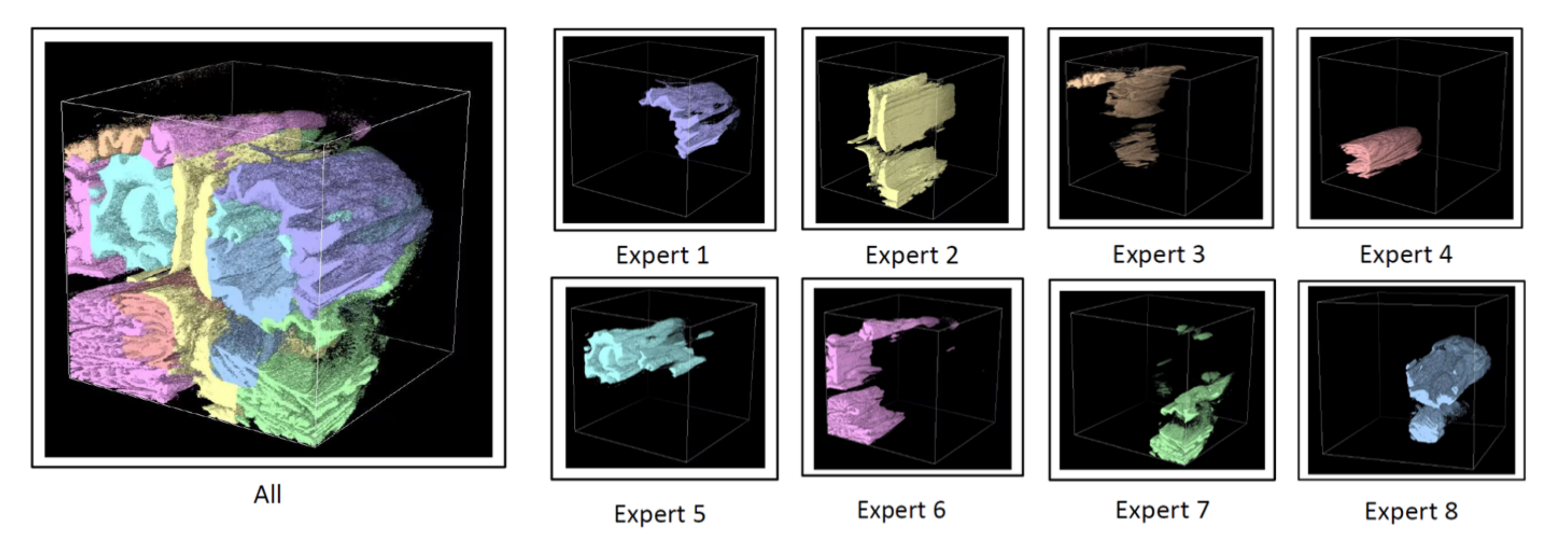}
\caption{The visualization of our scenes decomposition. It can be seen that our approach MoEC roughly decomposes a brain CT image into different parts of brain, cerebellum, background, and dispatches them to different experts.}
\vspace{-3mm}
\label{fig:semantic.decomposision}
\end{figure*}

Massive and growing visual data from various fields such as entertainment, surveillance, and computational imaging are generated nowadays, resulting in the urgent need for effective compression techniques for data storage, sharing, and transmission. Among different kinds of visual data, image and video compression have been well studied for many years and a large amount of codec standards have been proposed for commercial usage. However, the compression techniques of 3D/4D data like biomedical data are still under-explored to the best of our knowledge. Implicit Neural Representation (INR) has gained increasing attention from the community and is widely used in static view synthesis \cite{mildenhall2021nerf,muller2022instant,fridovich2022plenoxels,yu2021plenoctrees,sun2022direct}, dynamic view synthesis \cite{pumarola2021d,park2021hypernerf}, shape reconstruction \cite{wang2021neus,wu2022voxurf}, and SLAM \cite{zhu2022nice,rosinol2022nerf}. By overfitting DNNs to a complex scene, INR can represent the scene at high fidelity with the compact parameters of DNNs. This property makes INR serve as a promising data compression technique.

Although INR is promising for data compression, it has intrinsically limited spectrum coverage and needs to partition a complex scene into several local regions. Therefore, the compression performance heavily depends on the partition scheme. SCI \cite{yang2022sci} and TINC \cite{yang2022tinc} are two pioneering works that use INR to compress complex biomedical data. To be more specific, SCI analyzes the representation fidelity of INR from the spectrum perspective and develops an adaptive partition scheme to divide the scene into blocks with similar spectrum coverage. They then fit an INR into each divided block and serialize the parameters of INRs to compress the data. To achieve a higher compression ratio, TINC presents a tree-structured partition scheme to organize the parameters of INRs under a hierarchical architecture. Both SCI and TINC greatly improve the capability of INR and outperform commercial compression tools. However, both block-wise and tree-structured schemes are manually designed, which fail to learn the optimal partition and INRs jointly. Since manually developing the optimal partition scheme for a complex scene is very challenging, 

Taking consideration of the above factors, we have developed a novel Implicit Neural Representation (INR) compression technique known as Mixture of Experts Compression (MoEC). This method is capable of adaptively identifying the optimal partitioning scheme in an end-to-end manner, without the need for any data prior or manually designed scheme. Consequently, MoEC is capable of producing stable results that exhibit enhanced robustness with respect to the compression ratio. Our core design consists of two key components: an intelligent router network and expert network. Drawing inspiration from the switch-nerf\cite{mildenhall2021nerf}, we have incorporated Tutel \cite{hwang2022tutel} to facilitate the stable training of MoEC and the deployment of the router network.
For the expert network, we have employed SIREN. This is a departure from traditional INR methods that utilize Multilayer Perceptron (MLP) and Rectified Linear Unit (ReLU). Instead, SIREN uses a sine function as its activation function. This unique design enhances the expert’s sensitivity to frequency, enabling it to capture more high-frequency detail. Also we've found the compression process can achieve better performance on feature level, so we design an Encoder-Decoder Module as the head and tail of our model.

Our main contributions are summarized as follows:

\begin{itemize}
\item We propose MoEC, a novel mixture of experts compression method that can jointly learn the scene partition and INR experts in an end-to-end manner.

\item We carefully design the training and testing of MoEC to achieve efficient and stable implementation.

\item We conduct detailed experiments to demonstrate the advantages of our method against the existing approaches.
\end{itemize}

\section{Related Work}
\label{sec:relativework}
\noindent{\bf Data Compression\quad}
Data compression, which aims to efficiently store and transmit data, has recently become a major research focus of computer vision in the big data era. Before Deep Neural Networks (DNNs) become prevailing, many handcrafted image compression methods have been proposed and achieved promising results, such as JPEG \cite{pennebaker1992jpeg} and JPEG2000 \cite{rabbani2002overview}. Borrowing principles from image compression techniques, traditional video compression approaches like H.264 \cite{wiegand2003overview} and HEVC \cite{sullivan2012overview}, are designed to be both fast and accurate. More recently, DNN-based data compression methods have been becoming popular. Despite their success, these data compression techniques still lack the ability to deal with higher-dimension data such as biomedical volume. To handle the challenges, SCI \cite{yang2022sci} and TINC \cite{yang2022tinc} encode the intrinsic data structure with INRs, which turn out to work well on such inherently spatially structured data. However, combined with handcrafted rules, INR's representation ability on complex data suffers a serious limitation. In contrast to SCI and TINC, our approach leverages mixture of experts to find a suitable partition and raise the compression ratio significantly.

\noindent{\bf Implicit Neural Representation\quad}
As a novel way to represent continuous signals, implicit neural representation has been introduced to the field of data compression with promising results. For images, the INR function maps 2D coordinates to RGB values \cite{strumpler2022implicit}. For videos, the INR function maps time and 2D coordinates to RGB values \cite{chen2021nerv,zhang2021implicit}. For a 3D shape, the INR function maps the 3D coordinates to the values \cite{yang2022sci,yang2022tinc}. However, since INR has intrinsically limited spectrum coverage, therefore, existing methods need to partition a complex scene into several local regions. The compression performance heavily depends on the partition scheme.

\noindent{\bf Mixture of Experts (MoE)\quad}
The establishment of the original MoE is mainly based on the divide-and-conquer principle, in which it combines a vanilla Top-k gating layer to dispatch samples to k experts \cite{jacobs1991adaptive}. The MoE models are based on various expert models such as support vector machines \cite{collobert2001parallel}, Gaussian processes \cite{tresp2000mixtures}, or hidden Markov models \cite{fruhwirth2006finite}. With the rise of deep learning, many proposed methods in this field adopt neural network experts. Most recently, in order to adaptively find scene decomposition rules for large-scale scenes, Switch-NeRF \cite{zhenxingswitch} combines MoE with Neural Radiance Fields (NeRF) \cite{mildenhall2021nerf}. During the training process, they propose an auxiliary loss to balance the training of different experts. Instead, our method combines MoE with INRs to achieve higher ratios for neural data compression.
\section{Method}
\label{gen_inst}
\begin{figure*}[htbp]
\centering
\includegraphics[width=1\textwidth, height=0.48\textwidth]{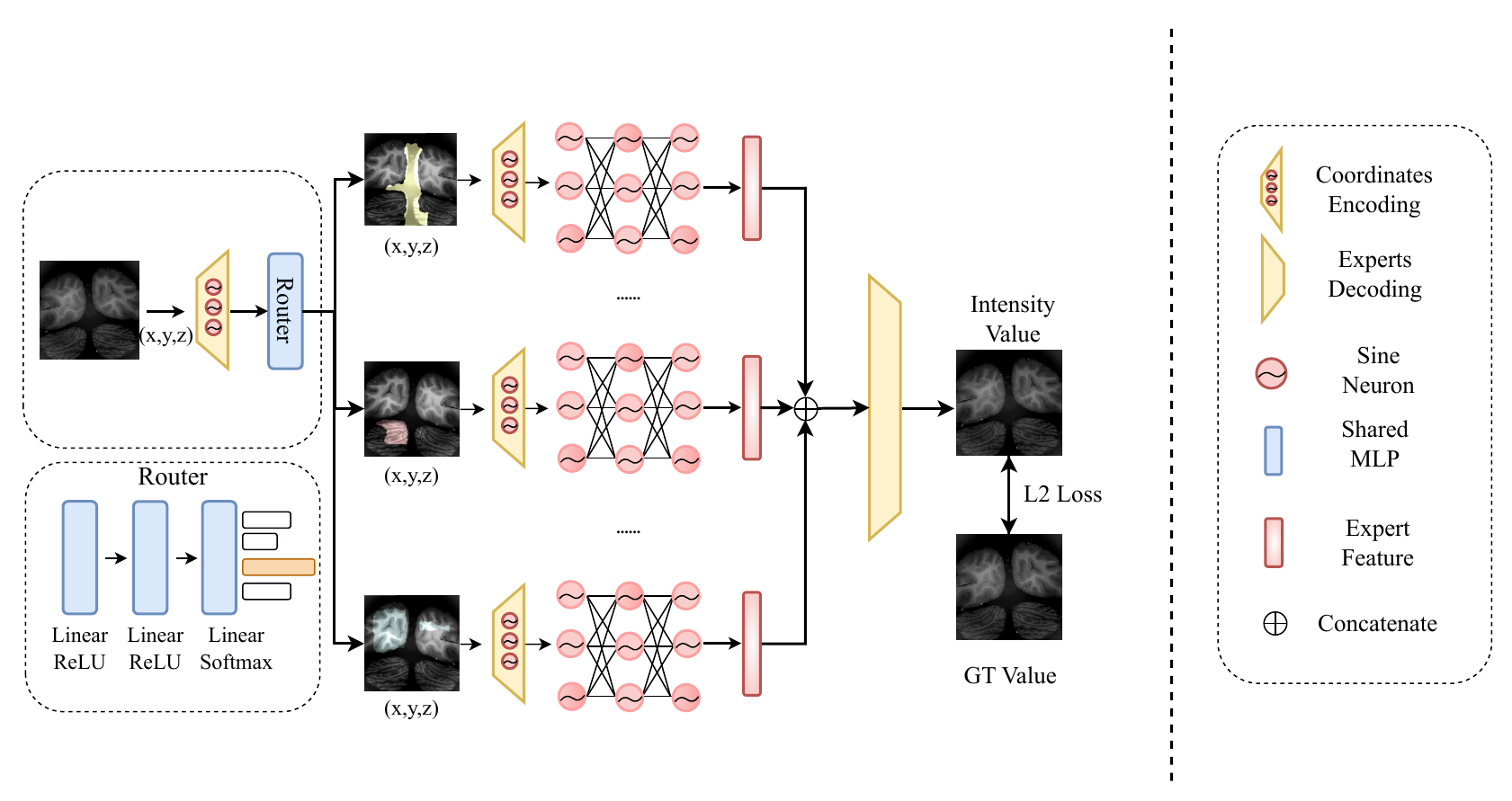}
\caption{The scheme of proposed approach MoEC. Sampled coordinates \textbf{(x,y,z)} first go through an intelligent gating network (router) and be dispatched to a favored expert for compression. Each specialized expert compresses different semantic parts of the scenes such as organs, and background.}
\vspace{-3mm}
\label{fig:pipe}
\end{figure*}

Based on the analyses conducted by SCI\cite{yang2022sci} and TINC\cite{yang2022tinc}, it has been observed that when it comes to describing diverse and complex data, INR (Implicit Neural Representation) experiences fidelity decay. This means that an INR network excels in responding effectively to specific frequency components of the data, but struggles to maintain fidelity across the entire spectrum. 
To address this issue, both SCI and TINC have attempted to mitigate the problem by manually dividing the entire scene into multiple cubes and subsequently reconstructing each segment individually. However, it's worth noting that these existing solutions have certain limitations. They do not always yield the optimal decomposition results, and they struggle to facilitate the reuse of partitioning strategies across different biomedical datasets due to variations in the frequency diversity of the data.
Hence, we introduce the concept of Mixtrue of Experts (MoE) into our data processing pipeline. In this article, we present our innovative solution, MoEC, which has been designed as an end-to-end framework. This approach decomposes the target data scenes into several subsets and dispatches them to specialized sub-models called experts for compression. A schematic of our MoEC can be seen in Fig. \ref{fig:pipe}. To this end, we first overview our method in Sec. \ref{sec:3.1}. Subsequently, we describe how routing network get the learning-based partition in Sec. \ref{sec:3.2}. Then, we introduce the expert MLP we use with sinusoidal activation in Sec. \ref{sec:3.3}. Furthermore, in Sec. \ref{sec:3.4}, we delve into the shared encoder and decoder components, which play a pivotal role in ensuring seamless communication between the experts and facilitating the reconstruction process. Finally, in Sec. \ref{sec:3.5}, we introduce the training strategy of our MoEC.

\subsection{Overview of Mixture of Experts Compression}
\label{sec:3.1} 
Our MoEC framework is primarily composed of several essential components: (1) an intelligent router denoted as $\mathbf{G}$, (2) a set of $\mathbf{n}$ experts represented as $\mathbf{(E_{i})_{i=1}^{n}}$, where each expert is a fully connected MLP utilizing the sine activation function, and (3) finally a unified encoder-decoder pair denoted as $\mathbfcal{E}$ and $\mathbfcal{D}$.

When given biomedical data in three or four dimensions, we can represent the grid coordinates of the data as $\mathbb{V}:=[-1,1]^{H \times W \times D (\times C)}$, and the mapping of the intensity value at coordinate $x$ can be denoted as $\mathbf{I}(x)$, where $x \in \mathbb{V}$.

First, our approach begins by performing a comprehensive full volume random sampling on $\mathbb{V}$ as batches. Because we found that random sampling in the entire volume allows MoEC to learn about the global distribution and knowledge in the initial stage compared to sequential sampling, thereby letting the entire system converge faster. 

Subsequently, the sampled points, denoted as $x$, are directed to the coordinates encoder $\textbf{E}(x)$ to transform the coordinates into high-dimensional features. Following the encoding stage, the router takes on the task of routing these points to specific top-k experts, ensuring that each point is directed to the most appropriate expert. For each point, an output is generated by the respective expert. These output features, which are processed through the softmax values of the router, are then passed to the unified decoder. The decoder's role is to reconstruct the final scene intensity, thereby completing the transformation of the input data into a coherent output. The whole process can also be described in the following Eq. \ref{eq:1}.
 
\begin{equation}
\label{eq:1}
\begin{aligned}
    f(\cdot) &= \sum_{i=1}^{n}  \mathbf{G}(\cdot)_{i}  \mathbf{E}_{i}(\cdot)  \\
   \mathbf{F}_{\bm{\theta}}(x) &=  \mathbfcal{D} \Big[ f \big( \mathbfcal{E}(x)  \big)   \Big]\\
\end{aligned}
\end{equation}
with $x$ a input coordinate point, $\mathbf{G(\cdot)}$ the gating network, $\mathbf{E(\cdot)}$ the experts, $\mathbfcal{E(\cdot)}$ the encoder, $\mathbfcal{D(\cdot)}$ the decoder, and $\bm{\theta}$ denoted those trainable parameters of all modules. The goal of our compression task is to minimize the difference between expected input data $\mathbf{I}(x)$ and decompressed data $\mathbf{F}_{\bm{\theta}}(x)$, so the loss function can be defined as Eq. \ref{eq:loss1}.

\begin{equation}
\label{eq:loss1}
    \mathcal{L}_{d} = \arg\min_{\bm{\theta}} \sum_{x \in \mathbb{V}} \Big[ \mathbf{I}(x) - \mathbf{F}_{\bm{\theta}} (x) \Big]^2
\end{equation}

Under this elaborate design of our pipeline, the loss can be passed to both the router and expert network during backpropagation, hence the whole network could be trained and optimized jointly. After enough training process, the router is able to directly learn an optimal division route towards the target data and boost a better compression quality. Details of each import module in our structure will be discussed in the following sections.

\subsection{Sparse Gating for Partition}
\label{sec:3.2}
As mentioned above, INR's spectral convergence makes it incapable of representing data with a broad spectrum. Therefore, a well-designed partition scheme is the critical factor to perform high-quality compression. Previous exploration mainly focus on hand-craft partition schemes, although they delivered a successful result and boost the compression ratio, however, these kinds of hand-craft ways are not efficiency friendly and not easy enough to reuse the partition result because the partition route heavily depends on the characteristic of target data. Instead, our learning-based partition scheme can be trained and optimized to find the optimal partition route in an end-to-end manner, in which target data with a broad spectrum will be decomposed into different semantic parts of the scenes, and these learned weak semantic scenes with different frequencies can be easier described by a single and simple INR expert.

In our work, we deepened the router based on the commonly used form of MoE layer proposed by Shazeer\cite{shazeer2017outrageously}, whose gating network consists of only one MLP layer. We have found that this kind of shallow MoE layer performs well in the previous work mainly because it is always set as in the relatively deep layer so that MoE can deal with high-level features to achieve a better routing. Inspired by the above analysis, we add another two linear layers before the gating network, as shown in Fig. \ref{fig:pipe}.

\subsection{Expert Compression}
\label{sec:3.3}
The primary role of the expert network is to compress each sub-scene determined by the router, and the expert network occupied for the majority of parameters in our compression results. Each of our compression experts follows the common 7 of Fully-Connected layers and selects sine as our activation function. As Sitzmann's analysis\cite{siren}, this type of MLPs are continuous and infinitely differentiable, making them exceptionally smooth. This smoothness can be advantageous in various applications. Besides, the sine function is periodic, which means that it can capture periodic patterns in data without introducing complex transformations like positional encoding in NeRF.

Under the premise of a limited number of parameters in our compression task, we trade a wide network for a deep one.  As long as the depth of our expert network is fixed, the width is directly determined by the total amount of network parameters, which is decided by the size and compression ratio of the target data size.

\subsection{Shared Encoder and Decoder}
\label{sec:3.4}
The shared encoder in our MoEC framework is responsible for encoding the input data and extracting meaningful representations that are conducive to compression. This encoder is a central component that takes the input data and produces a compact representation that can be effectively utilized by the individual experts. The shared encoder architecture is carefully designed to strike a balance between capturing essential information from the input scenes and providing a compressed representation that is amenable to further processing by the experts.

After input coordinates are assigned and processed by the specified expert, instead of using directly the output of the expert as the final prediction, we implement a decoder shared by all the experts and a gating network for the final voxel intensity value prediction. Introducing such a unified decoder enables experts to share information, each expert can make predictions not only based on the assigned data but also refer to the information of the entire data volume. Besides, the information on high-level features generated by deep expert can also be passed to the gating network.

The shared encoder and decoder components are essential for maintaining consistency and coherence in the data processing pipeline. They enable efficient communication between the experts and ensure that the final reconstructed output aligns with the input data. Additionally, the shared architecture allows for a more streamlined training process, as it promotes collaboration among the experts while maintaining a cohesive end-to-end framework.

\subsection{MoEC Training}
\label{sec:3.5}
One of the significant challenges encountered when training an MoE network is the issue of self-reinforcement. Eigen et al.\cite{eigen2013learning} highlighted this observation, noting that during the training process of an MoE model, experts that perform well on the initial samples tend to receive larger gate function outputs. Consequently, these experts are selected more frequently and undergo more extensive optimization. As a result, only a subset of experts in the MoE model becomes fully optimized, while the remaining experts do not receive adequate training. This phenomenon is commonly referred to as the load imbalance problem, which manifests as an issue in various MoE applications within the context of deep learning. This poses a challenge for our compression task as we are utilizing weights across multiple experts without effectively harnessing their potential, especially in our compression task with a limited number of parameters.

\noindent{\bf Balancing Loss \quad} To solve this problem, we incorporate a regularization loss denoted as $\mathcal{L}_{b}$ into our model, in accordance with the design principles outlined in \cite{fedus2022switch} for the differentiable load balance loss. This balancing loss serves the purpose of mitigating expert bias and effectively harnessing all parameters within our model. To elaborate, considering a scenario with $\mathbf{n}$ experts and a batch of $B$ input coordinates, where $\mathbf{c{i}}$ represents the count of coordinates assigned to the $i$-th expert ($\mathbf{E}_{i}$), the computation of the balancing loss is as following Eq. \ref{eq:bloss}:

\begin{equation}
\label{eq:bloss}
    \mathcal{L}_{b}=\frac{n}{B^2} \times \sum_{i=1}^{n} \mathbf{c}_{i} \sum_{x \in B} \mathbf{G}(x)_i
\end{equation}

In Eq. \ref{eq:bloss}, when we encounter a uniformly distributed input batch, both the number of assigned coordinates $\mathbf{c}_{i}$ and the gating value $\sum_{x \in B} \mathbf{G}(x)_i$ will converge to the same value, which is $\frac{n}{B}$. In such a scenario, the balancing loss $\mathcal{L}_{b}$ tends to approach the lower bound of 1.

Therefore, the total loss $\mathcal{L}$ can be rewritten as two losses:

\begin{equation}
    \mathcal{L}_{all} = \mathcal{L}_{d} +\lambda \mathcal{L}_{b}
\end{equation}
where $\lambda$ is a hyperparameter of balancing loss.

\noindent{\bf Expert Capacity \quad}
In addition to directly constraining the gate function, another idea is to ensure that each expert can be assigned an equal amount of training samples by the gate function. This method is generally used together with the regularization method. We are inspired by the idea from GShard \cite{lepikhin2020gshard}, which proposed to set an expert capacity $\mathbf{C}_f$ for each expert. Each token in the input sample is assigned to at most two experts. When the token assigned to an expert exceeds $\mathbf{C}_f$, the token will be discarded, which ensures that each expert Will not be over-optimized.

\noindent{\bf Balancing Dispatch\quad}
Furthermore, in our MoEC, training the gating network is the most challenging and time-consuming task. Therefore, an efficient and effective gating dispatch method will greatly reduce the training cost of our model. Under this consideration, we make use of uniform dispatch that was implemented by Tutel\cite{hwang2022tutel}. Given a batch $B$ coordinates, then in every on-one iteration, each expert can obtain at most $\frac{ \mathbf{C}_{f} B}{n}$ points, overflow points will be dropped. On the contrary, if experts obtain fewer sufficient points, they will be zero-padded. The introduction of uniform dispatch not only balances the weight of each expert, so that they will be trained in every iteration, reducing the difficulty of MoE training, but also improving computation and communication efficiency. 

\section{Experiments}
In this section, we initially carry out a quantitative comparison between our method and the current state-of-the-art (SOTA) methods, which include INR-based, data-driven, and commercial codec approaches. This comparison is conducted at both low and high compression ratios, utilizing two widely accepted metrics: PSNR and SSIM. For the qualitative analysis, we illustrate the efficacy of our method in contrast to the highest-performing baselines, specifically TINC and HEVC. TINC is a INR-based, with a manually-designed divide scheme method, whereas HEVC represents a prevalent commercial codec method.

\subsection{Experimental Settings}
\noindent{\bf Dataset\quad}
We utilized medical data obtained through HiP-CT, a technique that enables cellular-level imaging across multiple anatomical levels in various organisms. To verify the effectiveness of our model, we conducted extensive experiments on different organs, including the \textit{Lung}, \textit{Heart}, \textit{Kidney} and \textit{Brain}. For fair comparison and computation restriction, we crop the data into the same size of $\mathbf{256} \times \mathbf{256} \times \mathbf{256}$ voxels.

\noindent{\bf Metric\quad}
To quantitatively evaluate the performance of our model, we use two normal evaluation metrics: PSNR and SSIM. PSNR is positively correlated with Mean Square Error(MSE) and measures the degree of similarity between data, in a pixel-wise manner. SSIM, on the other hand, measures the similarity of data structures. 

\noindent{\bf Baseline Methods\quad}
The proposed methodology is compared comprehensively with current state-of-the-art techniques and organized into three categories: (i) widely available commercial compression methods such as JPEG\cite{wallace1992jpeg}, H.264\cite{wiegand2003overview}, and HEVC\cite{sullivan2012overview}; (ii) recently-developed data-driven deep learning-based ones, including DVC \cite{lu2019dvc}, and SSF\cite{agustsson2020scale}; (iii) INR-based ones, such as SIREN\cite{siren}, NeRF\cite{mildenhall2021nerf}, NeRV\cite{chen2021nerv} and TINC\cite{yang2022tinc}. The performance of all methods was evaluated at comparable compression ratios. It’s noteworthy that only INR-based techniques possess the capability to precisely regulate compression ratios, so we employed a distortion curve as a tool to illustrate and provide a comparative analysis of these methods. Performance was evaluated using the same metrics for all methods. 

\noindent{\bf Implement details\quad}
We utilize 2 sub-experts with \textbf{Top-1} gating. Each expert is made up of a 5-layer MLP with Sine activation and culminating in a final linear mapping layer. The gating network was designed with two layers of MLP, augmented with a ReLU layer, and a linear layer supplemented with softmax. The feature sizes of both the expert and gating were kept consistent, ultimately determined by the data size and compression ratio. Our methodology involved setting the sample batch to include 200,000 coordinates, with a process of 80,000 training epochs. We implemented a learning rate of $\mathbf{5} \times \mathbf{10}^\mathbf{-4}$, accompanied by an exponential decay scheduler. The optimization process was facilitated using the conventional Adam optimizer. Furthermore, acknowledging the disparity in convergence speed between the gating and expert, we adopted a phased training approach to circumvent potential adversarial scenarios during the training process. Upon reaching a predetermined step, we proceeded to freeze the gating network, thereby ceasing further updates.

\begin{figure*}[ht]
  \begin{minipage}{0.45\textwidth}
    \centering
    \includegraphics[width=0.75\textwidth, height=0.95\textwidth]{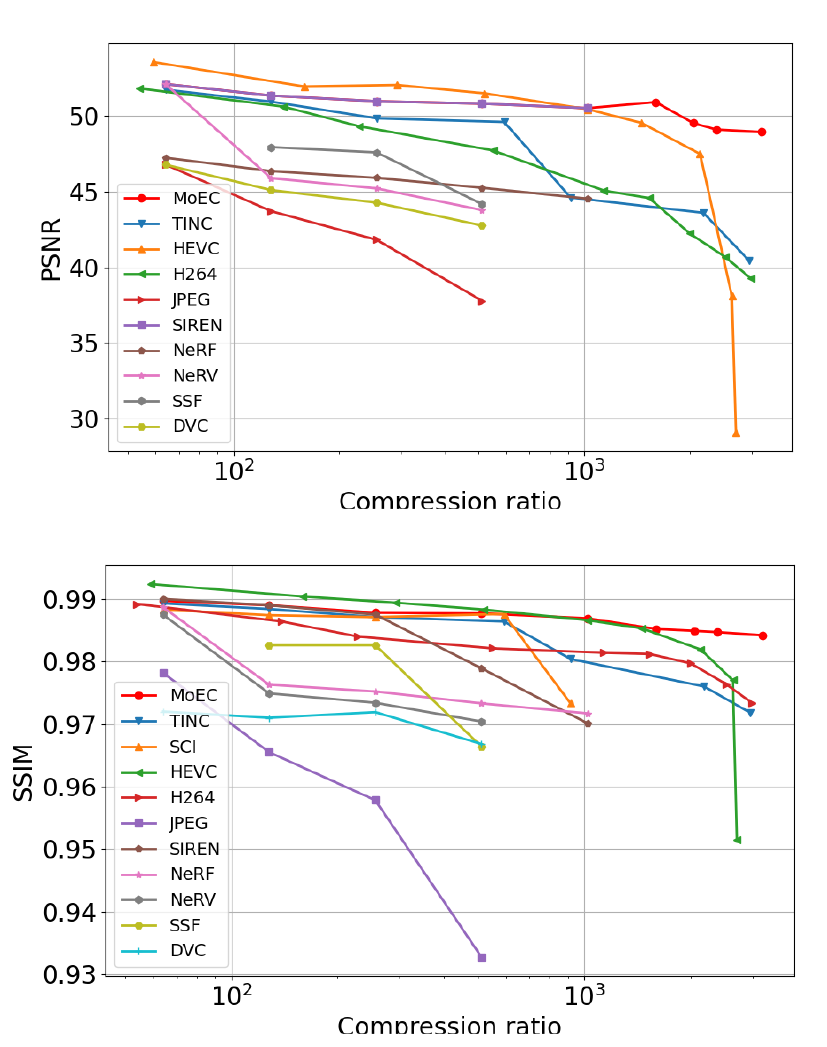} 
    \caption{Comparison of different compression methods on medical data at different compression ratios.}
    \label{fig:curve}
  \end{minipage}%
  \begin{minipage}{0.55\textwidth}
    \centering
    \renewcommand{\arraystretch}{1.3}
\centering
\resizebox{1\textwidth}{!}{
\begin{tabular}{lllllllll}

\toprule
           & \multicolumn{2}{c}{Lung} & \multicolumn{2}{c}{Heart} & \multicolumn{2}{c}{Kidney} & \multicolumn{2}{c}{Brain} \\ \cline{2-9} 
Methods    & PSNR        & SSIM       & PSNR        & SSIM        & PSNR         & SSIM        & PSNR        & SSIM     
\\ \midrule
SCI\cite{yang2022sci}        &45.99     &0.9769  &55.14      &0.9938    &46.01    & 0.9725      &49.02    &0.9899  

\\ \hline
TINC\cite{yang2022tinc}      &45.30      &0.9754    &56.26   &0.9984   &46.06   &0.9789      &49.57     &0.9924     

\\ \hline
SIREN\cite{siren}      &43.96     &0.9598    & 53.14    &    0.9971   &47.03   & 0.9769      &48.47     & 0.9877      

\\ \hline
NeRF\cite{mildenhall2021nerf}       &42.81    &0.9522   &49.20     &0.9924     &46.88      &0.9800     &47.00    &0.9819       

\\ \hline
NeRV\cite{chen2021nerv}       &42.95     &0.9540    &49.86      &0.9942     &45.95       &0.978    &45.01    &0.9753  

\\ \hline
JPEG\cite{pennebaker1992jpeg}       &41.85    &0.9650   &46.68  &0.9869     &43.96      &0.9690     &42.32     &0.9650         

\\ \hline
H.264\cite{wiegand2003overview}      &44.03    &0.9630    &53.16      &0.9963      &47.06    &0.9782      &51.32    &0.9936     

\\ \hline
HEVC\cite{sullivan2012overview}       &\textcolor{red}{47.27}     &\textcolor{red}{0.9829}    &\textcolor{red}{60.24}   &\textcolor{blue}{0.9991}      &\textcolor{red}{47.66}    &\textcolor{red}{0.9829}    &\textcolor{red}{52.61}    &\textcolor{red}{0.9947}          

\\ \hline
DVC\cite{lu2019dvc}        &42.47     &0.9531   &46.95     &0.9902    &44.76      &0.9723    &44.32    &0.9710          

\\ \hline
SSF\cite{agustsson2020scale}        &45.04     &0.9699  &51.14      &0.9953    &47.07    & 0.9719      &47.65    &0.9849       

\\ \hline
MoEC(Ours) &\textcolor{blue}{46.68}     &\textcolor{blue}{0.9789}      &\textcolor{blue}{58.59}       &\textcolor{red}{0.9992}     &\textcolor{blue}{47.13}      &\textcolor{blue}{0.9810}           &\textcolor{blue}{52.10}      &\textcolor{blue}{0.9942} 

\\ \bottomrule
\end{tabular}
}

    \captionof{table}{The average metrics on each dataset are presented as the compression ratio varies from 64 to 1024, with the \textcolor{red}{best} and \textcolor{blue}{second best} results highlighted by color.}
    \label{table:low}
  \end{minipage}
\end{figure*}

In our experiment, all input data are cropped into $\mathbf{256} \times \mathbf{256} \times \mathbf{256}$ and intensity values will be normalized to $0 \sim 100$ in the training. To verify the superiority and robustness of our method, we conducted experiments at various compression ratios, including 64x, 128x, 256x, 512x, and 1024x. As shown in Tab. \ref{table:low}, our MoEC outperformed other methods on almost all dataset.

\subsection{Evaluation on Low Compression Ratios} 

\subsubsection{Quantitative Evaluation}
Our approach was initially evaluated under low compression ratios ranging from 64x to 1024x, following the same setting as TINC. The results of these experiments are presented in Tab. \ref{table:low}. It is evident from the data that our method consistently outperforms current state-of-the-art techniques across all datasets, and produces comparable results with HEVC. The rate-distortion curve depicted in the Fig. \ref{fig:curve} provides a comparative view of the average performance of each method at varying compression ratios.

Notably, our method demonstrates significant superiority over manually designed partitioning methods such as TINC. This underscores the effectiveness of our learnable partitioning scheme.

Interestingly, our method yields results that are on par with the High Efficiency Video Coding (HEVC). Compression techniques like HEVC inherently hold an advantage over Implicit-Neural-Representation-based (INR-based) methods, particularly when dealing with medical data. These data often contain substantial redundant information and noise. INR methods, which tend to overfit the data using a neural network, include all redundant information and noise. This inevitably leads to a trade-off between fidelity and redundancy during optimization, resulting in some loss of real information — a problem not encountered with HEVC.

However, our method’s characteristic of retaining all source image information, thereby preventing information loss, offers a more reliable alternative to HEVC’s strategy of using inter-frame and intra-frame information to select the optimal compression strategy. As illustrated in Fig. \ref{fig:low}, on the first row of kidney, while HEVC lost some important frame information, our method managed to preserve it. Our observations indicate a frequent occurrence of this situation, specifically within P-frames. This further demonstrate the robustness and reliability of our methodology.

\subsubsection{Qualitative Evaluation}

\begin{figure}[t]
\centering
\includegraphics[width=0.43\textwidth, height=0.40\textwidth]{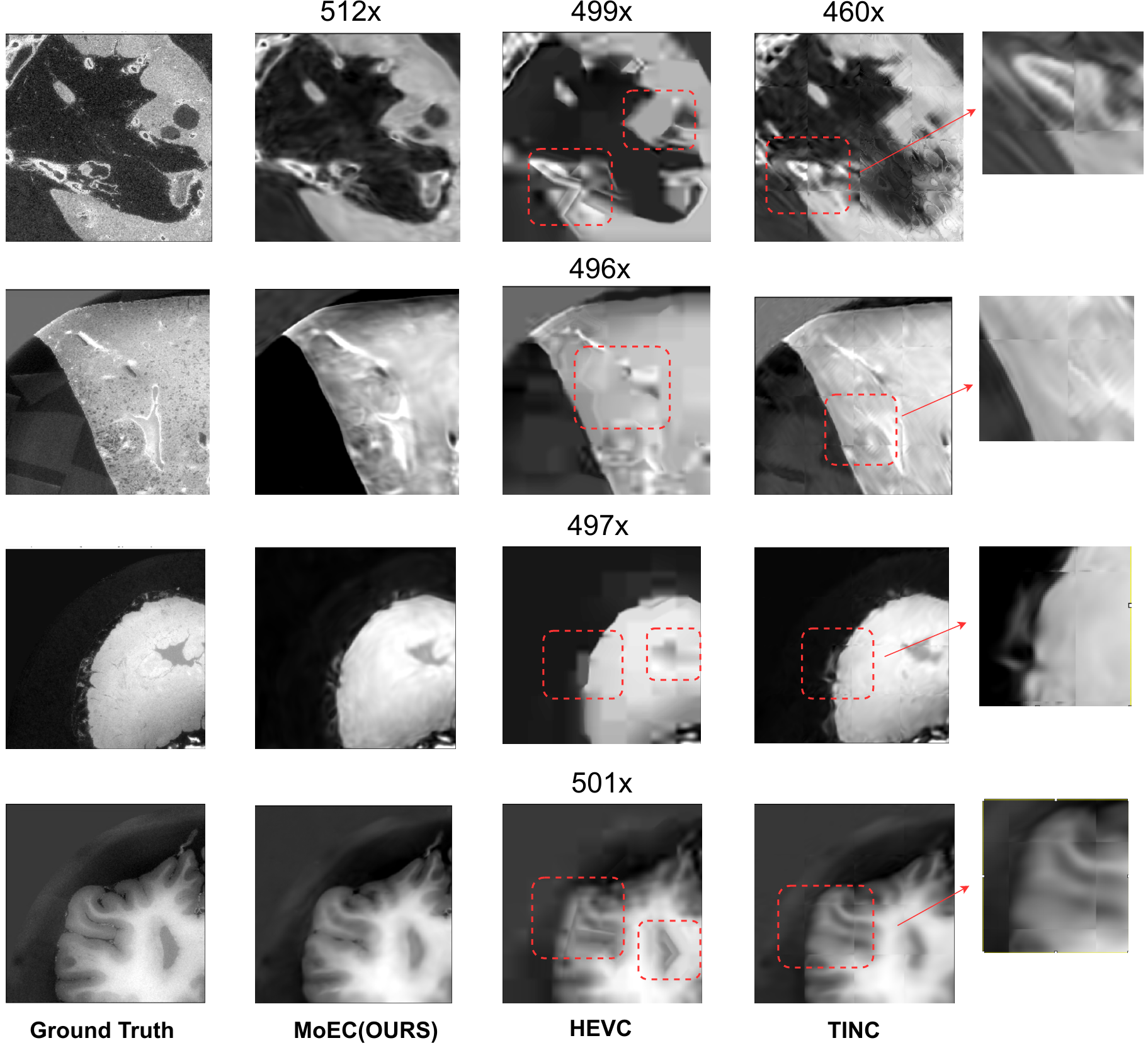}
\caption{Qualitative comparison with SOTA of decompressed medical data slice, under compression ration around 500$\times$. }
\vspace{-3mm}
\label{fig:low}
\end{figure}

We also provide a visual comparison. Some visual results of decompressed organ slices are reported in Fig. \ref{fig:low}. We consistently outperform current SOTA INR-based methods (TINC), and commercial compression methods (HEVC).

TINC employs an octree structure to partition a volume of $\mathbf{256} \times \mathbf{256} \times \mathbf{256}$  into 32,768 sub-blocks, each measuring $\mathbf{32} \times \mathbf{32} \times \mathbf{32}$. It then allocates shared parameters based on the spatial distance between these blocks. However, this intricate design causes costly effort for distance searching, not only results in a waste of compression time, but also leads to significant block artifacts between the blocks.

While HEVC excels at capturing details in “I Frames”, it struggles significantly with “P Frames”, leading to intolerable block artifacts. This could lead to a severe loss of information in the P frames, causing a spatial discontinuity in the data volume. These artifacts compromise the usability of the data for medical diagnoses and downstream tasks.

\subsection{Evaluation on High Compression Ratios} 
\subsubsection{Quantitative Evaluation}

With the advancement in the precision of medical equipment, we are now capable of performing diagnoses at the micron level. Concurrently, the volume of incoming medical data is escalating, with data sizes reaching into the thousands of gigabytes becoming increasingly commonplace. The need to transmit such colossal data necessitates a method capable of achieving an extraordinarily high compression ratio while preserving high fidelity at the same time. In light of this, we have conducted experiments on extreme compression ratios of 1000x, 1500x, 2000x, 2500x, 3000x, and 6000x. The results of these experiments are presented in Tab. \ref{table:high}. The results demonstrate that our method significantly outperforms other methods under high compression ratios. This includes both INR methods that are manually divided, such as TINC, as well as codec methods like H.264 and HEVC.

In instances of extreme compression ratios, the TINC parameter sharing design can lead to significant aliasing issues in adjacent blocks, particularly those containing both organ and background elements. And as previously noted, HEVC tends to lose information in P-frames. Consequently, when the compression ratio escalates rapidly, the number of P-frames also increases, resulting in a severely deterioration in performance.

Furthermore, the maximum compression ratio that HEVC can achieve across all datasets is approximately 2700x, and it is incapable of compressing at higher rates. Indeed, our experiments have shown that when the compression ratio surpasses 2000x, HEVC tends to perform poorly in reproduction, resulting in a substantial loss of information during the compression process. In contrast, our method manages to preserve the majority of the data information, even when the compression ratio exceeds 6000x.

\begin{figure*}
\centering
\begin{minipage}{.18\textwidth}
\centering
\begin{tabular}{llll}
    \toprule
         & PSNR     & SSIM     \\
    \midrule
    Top-2     &51.59      &0.9937   \\
    10k steps &50.45 &0.9936 & \\
    w/o loss    &51.29       &0.9934   & \\
    Experts-3    &\textbf{51.66}     &\textbf{0.9938} & \\
    Experts-4    &51.40   &0.9935  & \\
    
    \midrule
    \midrule
    MoEC    &51.56     &0.9931 & \\
    \bottomrule
\end{tabular}
\captionof{table}{Ablation study}
\label{table:ablatin}
\end{minipage}
\hfill 
\begin{minipage}{.7\textwidth}
\centering
\begin{tabular}{lllllllll}

\toprule
    & \multicolumn{2}{c}{MoEC(Ours)} & \multicolumn{2}{c}{TINC}  & \multicolumn{2}{c}{HEVC} & \multicolumn{2}{c}{H.264} \\ \cline{2-9} 
Methods    & PSNR        & SSIM       & PSNR        & SSIM        & PSNR         & SSIM        & PSNR        & SSIM         

\\ \midrule
1500$\times$   &\bf{49.76}  &\bf{0.9852}  
&46.11 & 0.9834 
&49.52   &0.9853     
&44.56     &0.9821        

\\ \hline
2000$\times$     &\bf{49.10}    &\bf{0.9843}  
&44.43      &0.9782    
&47.47   &0.9819   
&42.22    &0.9797     

\\ \hline
2500$\times$       &\bf{49.33}     &\bf{0.9847}    
&42.53   &0.9727  
&43.41    &0.9770  
&40.69    &0.9763   

\\ \hline
3000$\times$      &\bf{49.17}      &\bf{0.9842}    
&40.43     &0.9781  
&29.09   &0.9515   
&38.11   &0.9683

\\ \hline
6000$\times$      &\bf{48.16}    &\bf{0.9838}
&34.15  &0.9703   &x     &x   &x    &x        
\\ \bottomrule
\end{tabular}
\captionof{table}{The average metrics of each method on four organ datasets under high compression ratios. Mark 'x' means cannot be tested on that compression ratio.}
\label{table:high}
\end{minipage}
\end{figure*}
\subsubsection{Qualitative Evaluation}
\begin{figure}[t]
\centering
\includegraphics[width=0.43\textwidth, height=0.40\textwidth]{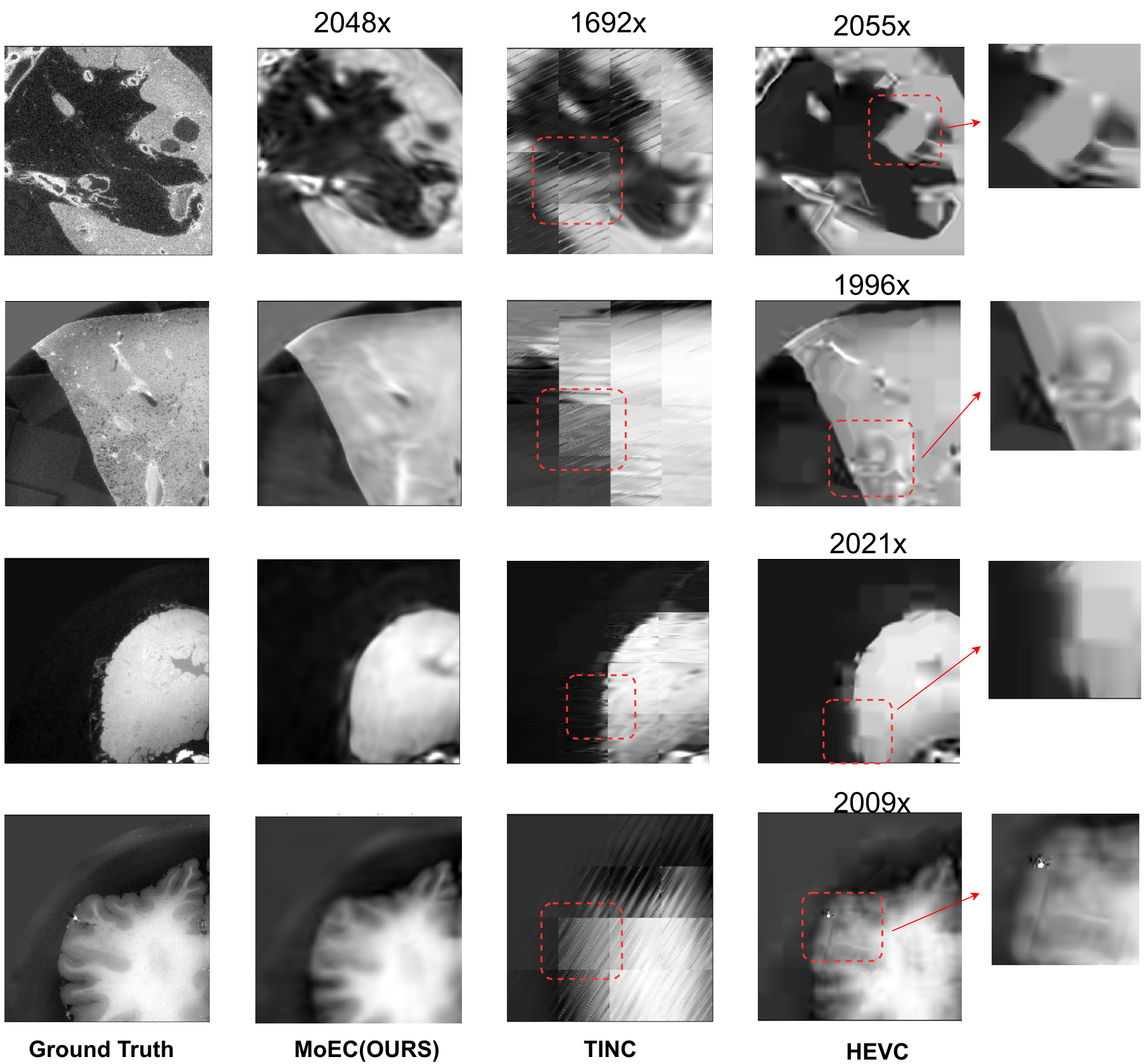}
\caption{Qualitative comparison with SOTA of decompressed medical data slice, under high compression ration. }
\vspace{-3mm}
\label{fig:high}
\end{figure}

The visual comparison showcased in Fig. \ref{fig:high} is particularly noteworthy. It's worth noting that our method stands out as the only one capable of maintaining high fidelity even when the rate exceeds 2000x. Under conditions of extremely high compression ratio, it’s evident that both TINC and HEVC generate significant artifacts, with the occurrence of bad frames also being observed. 

As one can observed that when the compression ration surpass 1500x, the block artifacts between TINC and HEVC becomes strikingly apparent. TINC (3-column) suffers from aliasing problem across all dataset, especially at the edges of the organ and the background. HEVC (3-column) produces a large number of 'bad frames' with highly blurred visual effect, mainly attribute to the frequent emergence of block artifacts compromises the spatial continuity of the data.

\subsection{Ablation Study}
Apart from the main experiment, We also conducted ablation experiments to analyze and verify the different modules in our model. The experiments conducted are configured to run for a total of 80,000 epochs. These experiments specifically focus on the analysis of a brain dataset and operate under a compression ratio of 512x.

\noindent{\bf Top-k Route\quad}
 We conducted experiments to evaluate the effect of using different \textbf{Top-k} values on the compression results. When using 2 experts, we tested \textbf{Top-1} and \textbf{Top-2}. The experimental results are shown in Tab. \ref{table:ablatin}, showing that using \textbf{Top-2} improves the accuracy of compression. However, \textbf{Top-2} also requires more training time and memory than \textbf{Top-1}. Thus, there is a trade-off between accuracy and efficiency. On the other hand, When operating under the same number of epochs, employing the \textbf{Top-2} strategy enables the model to achieve faster convergence, as depicted in Tab. \ref{table:ablatin}. Specifically, at the benchmark of 10,000 epochs, \textbf{Top-2} has already reached an acceptable PSNR, outperforming \textbf{Top-1}. Our analysis suggests that the use of \textbf{Top-2} allows the expert to encounter a broader range of input coordinates in a shorter time span, thereby mitigating the influence of gating dispatch bias on the training process.
 
\noindent{\bf Balancing Loss\quad}
Balancing Loss was introduced to address the issue of imbalance in MoE. Without regularization of balancing Loss, the parameters can not be fully utilized, resulting in reduced compression effectiveness. 
 
\noindent{\bf Expert Number\quad}
In our experiments, we’ve observed that the correlation between the number of experts and the model’s performance isn’t always positive. This is attributed to the necessity of judiciously distributing parameters across each module, given the constraints of a finite parameter budget. Consequently, an increase in the number of experts implies a reduction in the allocation of parameters per expert, which could potentially diminish the performance of individual experts. 

In the course of our experiment, we found that the scenes within the cropped images were relatively simple. This simplicity allowed us to achieve excellent results with the Top-1 gate by utilizing just two experts. As a result of this efficiency and effectiveness, we decided to proceed with this particular configuration.

\section{Conclusion and Limitation}
\noindent{\bf Limitation\quad} As a nascent technology, our work has room for improvement. In the future, we plan to implement more stringent supervision to regulate the gating network’s capacity to edit target data flexibly. This is a topic that warrants further exploration. 

\noindent{\bf Conclusion\quad}
In this work, we have proposed MoEC, first offering an end-to-end mixture-of-expert(MoE) framework in INR compression task. With elaborately design of our gating network, MoEC not only can achieve high-quality compression, but also learn to decompose the scenes simultaneously. Our methodology primarily draws upon INR and MoE, massive experiments demonstrate the effectiveness of our network. This is a groundbreaking study that combines the two, with the aim of providing inspiration for future research in these fields.

{
    \small
    \bibliographystyle{ieeenat_fullname}
    \bibliography{main}
}

\clearpage
\appendix
\maketitlesupplementary

\setcounter{table}{0} 
\setcounter{figure}{0}
\setcounter{section}{0}
\setcounter{equation}{0}
\renewcommand{\thetable}{A\arabic{table}}
\renewcommand{\thefigure}{A\arabic{figure}}
\renewcommand{\thesection}{A\arabic{section}}

\definecolor{cvprblue}{rgb}{0.21,0.49,0.74}

\section{Overview}

This supplementary material provides a detailed analysis of our methodology and expands upon our approach and experiments.  In Sec. \ref{comparison}, We have provided a comprehensive visual comparison, showcasing the effectiveness of our method in contrast to the state-of-the-art (SOTA) methods. Furthermore, to provide a more dynamic and detailed comparison, we have created a video that illustrates the process and results of data decompressed using these methods. In Sec. \ref{sec: Model Compression}, we try to use model compression methods to further compress the parameters of the neural network into smaller sizes. In Sec. \ref{expert},  we've verified the correlation between expert number and reconstruction fidelity. In Sec. \ref{time}, we conducted a comparative analysis of the time required for compression and decompression across different methods. In Sec. \ref{data-complex}, We've done a preliminary investigation into the spectrum bias characteristics inherent in the INR method.

\section{Visual comparison between MoEC and SOTA methods }
\label{comparison}

\begin{figure}[t]
\centering
\includegraphics[width=0.43\textwidth, height=0.40\textwidth]{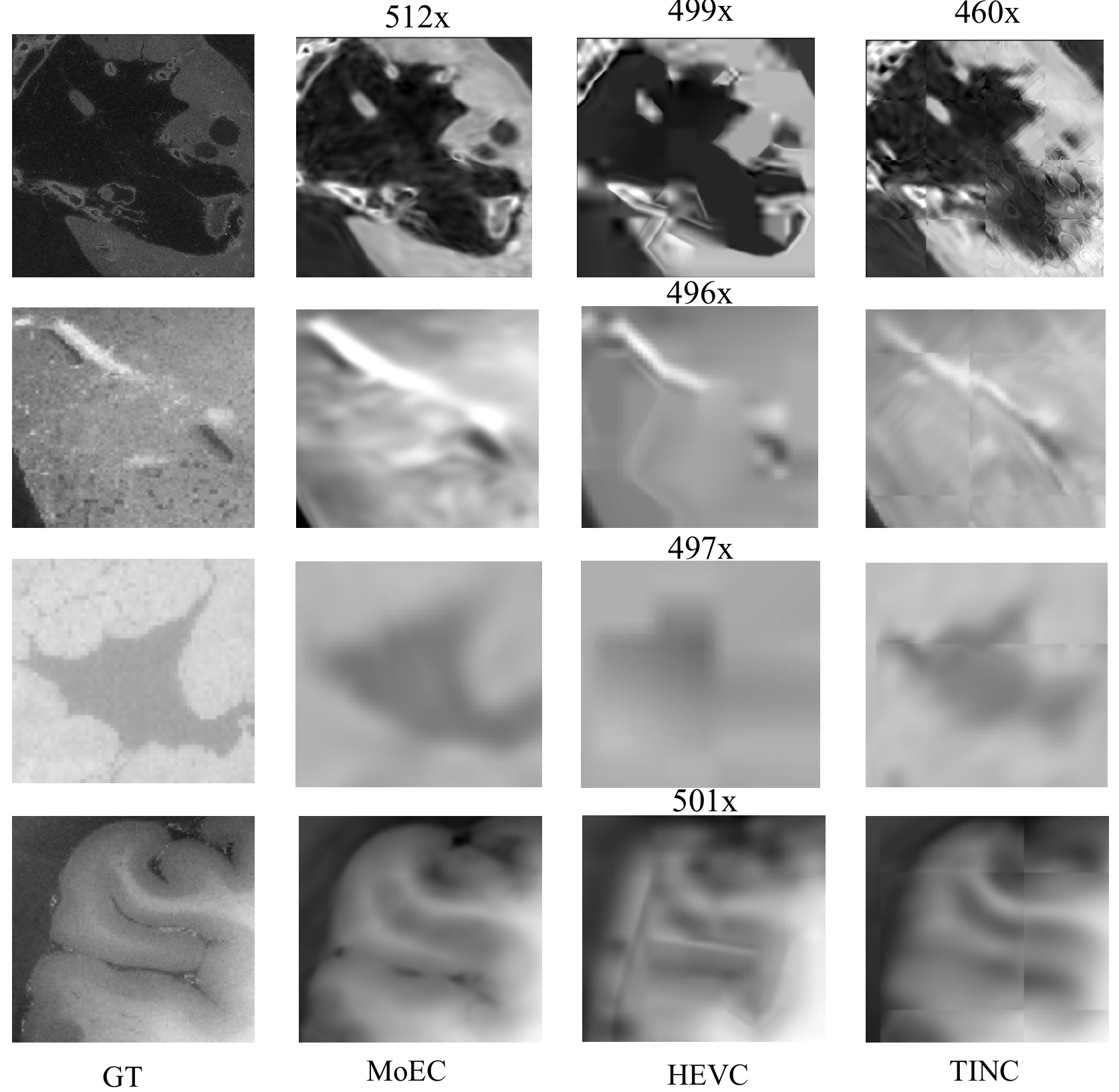}
\caption{Visual comparison with SOTA methods. Row 1-4 are the regions-of-interest (ROIs) on different medical data slice(Kidney, Lung, Heart, Brain, respectively). The compression ratios are labeled at top. }
\vspace{-3mm}
\label{fig:sup-low}
\end{figure}

As previously noted, the High Efficiency Video Coding (HEVC)\cite{sullivan2012overview} frequently experiences a high incidence of defective frames during data compression, with these anomalies predominantly manifesting in P-Frames. Fig. \ref{fig:sup-low} illustrates this issue, highlighting the severe block artifact introduced by HEVC, which obscures a significant portion of the data’s information. In contrast, our proposed method demonstrates consistent performance across all frames, successfully recovering the majority of the information. This not only attests to the effectiveness of our approach but also underscores its robustness.

\section{Toward Model Compression }
\label{sec: Model Compression}

This article introduces a method to compress a dense medical volume into neural network parameters. In fact, there has been a lot of work in the past dedicated to further lightweighting the parameters of neural networks, such as pruning\cite{molchanov2016pruning,frankle2018lottery}, distillation\cite{hinton2015distilling, beyer2022knowledge}, and quantization\cite{gholami2022survey, jacob2018quantization, nagel2021white}. This chapter attempts to use quantization and entropy coding techniques to further reduce the storage size of our model. It is worth mentioning that entropy coding is a coding mode based on information entropy and is lossless compression, while quantization is lossy compression.

\subsection{ Quantization }
Model quantization is a technique used to reduce the computational and memory requirements of machine learning models. It achieves this by reducing the precision of the numbers used to represent the model's parameters, typically from 32-bit floating-point numbers to lower-precision formats like 8-bit integers. The main advantage of quantization is that it can significantly reduce the computational and memory requirements of machine learning models, making them feasible to run on lower-resource devices like mobile phones or embedded devices. However, the trade-off is that quantization can lead to a drop in model accuracy, especially for models that are not trained with quantization in mind.

There are two main types of quantization: 1. \textbf{Weight Quantization}: This involves converting the weights of the neural network from floating-point representations to lower-precision representations. This can significantly reduce the model size and the computational requirements of the inference phase. 2. \textbf{Activation Quantization}: This involves quantizing the activations or the intermediate outputs between layers of the neural network during the inference phase. This can further reduce computational requirements and memory usage. Quantization can be performed either during training (quantization-aware training) or after training (post-training quantization). Post-training quantization(PTQ) is simpler and doesn't require retraining the model, but it may lead to a larger drop in accuracy because the model wasn't trained to cope with the reduced precision. Quantization-aware training(QAT) modifies the model to learn lower-precision weights during training, which often leads to better accuracy because the model learns to cope with the reduced precision.

\begin{table*}[htbp]
  \caption{Quantization Results. The original volume size is 32MB. }
  \label{table:quant}
  \centering
\begin{tabular}{l|lll|lll}
\toprule

 & \multicolumn{3}{c|}{MOEC} & \multicolumn{3}{c}{Q-MOEC} \\
Setting  & Size(KB)   & PSNR(dB)   & SSIM   & Size(KB)    & PSNR(dB)    & SSIM   \\
\midrule 
256x             &    128.0    &    51.95    &    0.9942    &    34.6     &    51.96     &    0.9937    \\
512x             &    69.1    &    51.19    &    0.9933    &    19.7     &     51.02    &     0.9933   \\
1024x            &    33.6    &    50.76    &    0.9929    &     9.6    &    48.98     &    0.9923   \\
\bottomrule
\end{tabular}
\end{table*}

In this subsection, we use the quantization library that comes with pytorch to perform symmetric PTQ on our model parameters. Unlike other work, we do not quantify the activation values, because activation quantization often causes a significant drop in model performance, and does not Will reduce more storage space. The results are shown in Tab. \ref{table:quant}.

\subsection{ Entropy Coding }
Huffman coding\cite{huffman1952method} is a widely recognized entropy encoding technique employed for lossless data compression. The fundamental principle of Huffman coding is to allocate shorter codes to characters that appear more frequently and longer codes to those that appear less frequently. The core concept underlying Huffman coding is a binary tree sorted by frequency, which is utilized to identify the most efficient representation for each data piece. Huffman's innovation was to transform a frequency-sorted binary tree into a table for fast encoding and decoding. In this binary tree, the leaf nodes represent the symbols (characters in the file being compressed), and their frequency of appearance dictates their position within the tree. The character that appears most frequently is situated at the root of the tree, while the least frequent character is positioned furthest from the root. As a result, characters that appear more frequently are assigned shorter, faster codes, while less frequent characters are given longer codes. This reduces the total number of bits required for the data, resulting in efficient compression. Huffman coding is extensively utilized in contemporary digital systems, including computer networks, fax machines, and modems, and as a subroutine in other data compression algorithms.

The following Pseudo-code \ref{pseudo:huff} outlines the Huffman Coding process. It accepts a set of characters along with their frequencies as input. It forms a priority queue from this set and then, in a loop, removes the two nodes with the lowest frequencies, generates a new parent node with its frequency being the sum of the two children's frequencies, and reinserts the new node into the queue. This process continues until only one node, the root of the Huffman tree, remains in the queue, which is then returned by the procedure.

\begin{algorithm}
\caption{Huffman Coding}
\label{pseudo:huff}
\begin{algorithmic}[1]
\Procedure{Huffman Coding}{$C$} \Comment{$C$ is a set of characters with their frequencies}
    \State $n \gets |C|$
    \State $Q \gets C$ \Comment{Initialize priority queue $Q$ with $C$}
    \For{$i = 1$ to $n-1$}
        \State allocate a new node $z$
        \State $z.left \gets x \gets \text{Extract-Min}(Q)$
        \State $z.right \gets y \gets \text{Extract-Min}(Q)$
        \State $z.freq \gets x.freq + y.freq$
        \State Insert $Q, z$
    \EndFor
    \State \Return Extract-Min(Q) \Comment{Return the root of the Huffman tree}
\EndProcedure
\end{algorithmic}
\end{algorithm}

\begin{table}[htbp]
  \caption{Entropy Coding Results. The first column indicates the size of the model prior to entropy encoding, the second column shows the size of the model following entropy encoding, and the third column gives the ratio of the model size before and after entropy encoding.}
  \label{table:huff}
  \centering
\begin{tabular}{l|lll}
\toprule
                 & Before                      & After        & Ratio                    \\ \midrule
256x             &  128.0 KB                         &  97.7 KB                     &  76.3\%                    \\
512x             &  69.1 KB                         &  64.3KB                     &   93.1\%                   \\
1024x            &  33.6 KB                         &  39.5 KB                     &  117.6\%                     \\ \bottomrule
\end{tabular} 
\end{table}

The results of Huffman Coding are shown in Tab. \ref{table:huff}. It can be observed that under the experimental setting of 1024x, the size of the model after entropy encoding does not decrease but increases instead. This is attributed to the fact that Huffman encoding necessitates the preservation of a Huffman tree. The size of this tree itself takes up storage space, hence the final size may not necessarily be smaller.

\section{Relationship between Reconstruction Fidelity and Expert Number }
\label{expert}

\begin{figure}[htbp]
\centering
\subfloat[Brain]{
    \label{Brain}
    \includegraphics[width=0.25\textwidth]{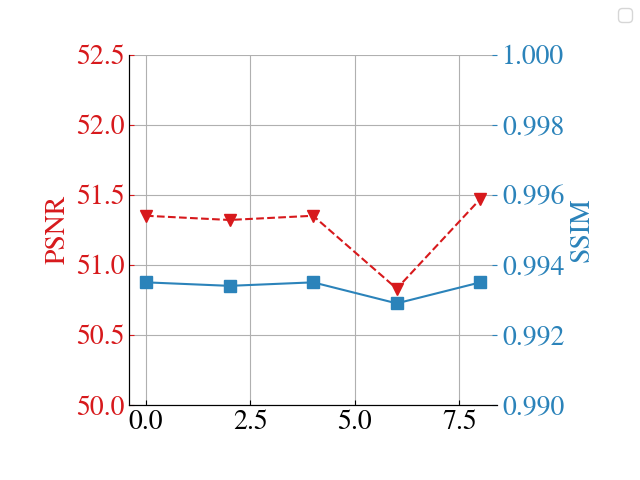}
}
\subfloat[Heart]{
    \label{Heart}
    \includegraphics[width=0.25\textwidth]{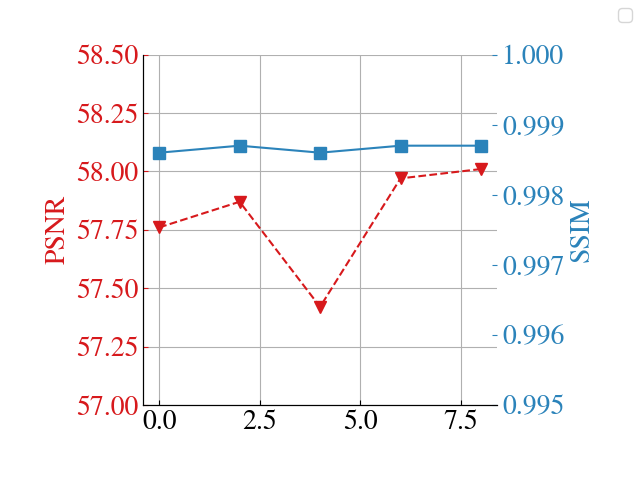}
}\\
\subfloat[Kidney]{
    \label{Kidney}
    \includegraphics[width=0.25\textwidth]{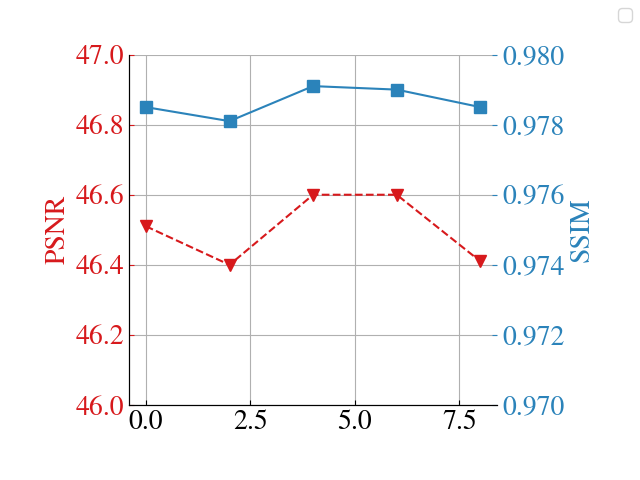}
}
\subfloat[Lung]{
    \label{Lung}
    \includegraphics[width=0.25\textwidth]{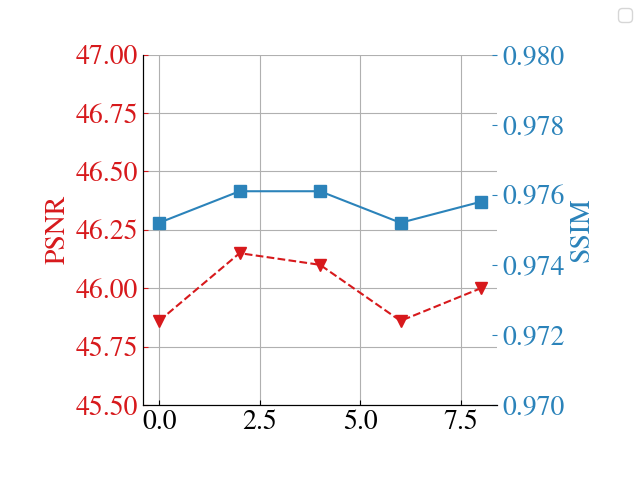}
}
\caption{Relationship between Expert Number and Reconstruction Fidelity. As one can observe that the quality of reconstruction remains consistent with the fluctuation in the number of experts, maintaining a PSNR variation within 1db and an SSIM variation within 0.01. And there isn't a  obvious positive proportional correlation.}
\label{sub-expert}
\end{figure}

The network’s ability to learn an optimal partitioning scheme is significantly influenced by the number of experts. It can be anticipated that choosing a different number of experts will result in a varied partitioning scheme, which in turn impacts the quality of the reconstruction. To further investigate the relationship between the number of experts and the reconstruction quality, we conducted additional validation tests. These tests were performed under a consistent compression ratio of 512x, with the number of experts varying from \textbf{2} to \textbf{10}. The performance was subsequently evaluated after a training period of 80,000 steps.
The Fig. \ref{sub-expert} illustrates the robustness of our method, as validated by the experiment. When the number of experts varies from \textbf{2} to \textbf{10}, the PSNR fluctuation remains within 1db, and the SSIM variation does not exceed 0.01. Despite these findings, the results do not exhibit a strong positive correlation. This is attributed to the fact that with a fixed parameter budget, an increase in the number of experts results in each individual expert network receiving fewer parameters. Consequently, the representational capacity of each expert network diminishes, presenting a genuine trade-off scenario. It's important to note that an increase in the number of experts escalates the complexity of training the router network, leading to slower convergence and necessitating a longer training duration. Therefore, maintaining a constant training time across all experiments in our study is unfavorable for networks with a higher number of experts.

\section{Compression and Decompression time}
\label{time}
\begin{table}
  \caption{Compression Efficiency}
  \label{table-time}
  \centering
  \begin{tabular}{lll}
    \toprule
         & Compression time     & Decompression time               \\
    \midrule
    H.264     &0.2487     &0.1245          \\
    HEVC     &0.6338    &0.1437               \\
    TINC    &13846.65      &0.2536                \\
    \midrule
    \midrule
    MoEC(ours)    &2649.00    &2.0300            \\
    \bottomrule
  \end{tabular}
\end{table}

We also conducted a verification of the efficiency of compression and decompression across different methods. Specifically, for the INR\cite{yang2022tinc} method, we employed identical hardware (a single Tesla v100), maintained a consistent batch size (200,000 coordinates per iteration), and logged the time taken for compression and decompression following a training duration of 80,000 steps.
Tab. \ref{table-time} clearly illustrates that the INR-based method, which employs a neural network to overfit the target data, requires a considerably longer compression time compared to the codec method. However, the time needed for INR inference is drastically reduced, thereby significantly closing the gap with the decompression time required by the codec method.

\section{INR's fidelity decay with spectrum concentration.}
\label{data-complex}

\begin{figure}[!htbp]
\centering
\subfloat{
    \includegraphics[width=0.24\textwidth]{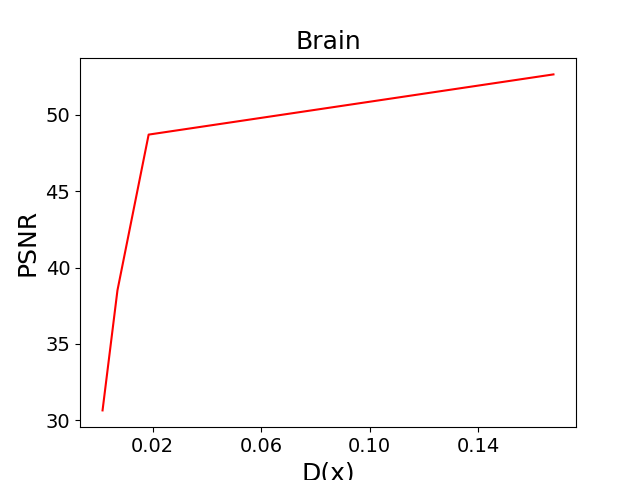}
}
\subfloat{
    \includegraphics[width=0.24\textwidth]{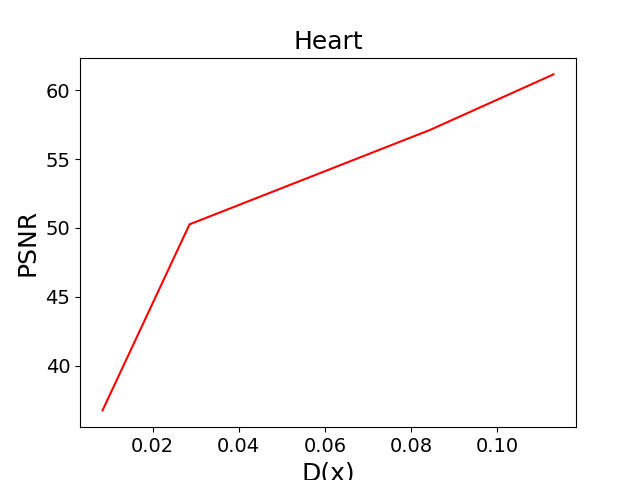}
}\\
\subfloat{
    \includegraphics[width=0.24\textwidth]{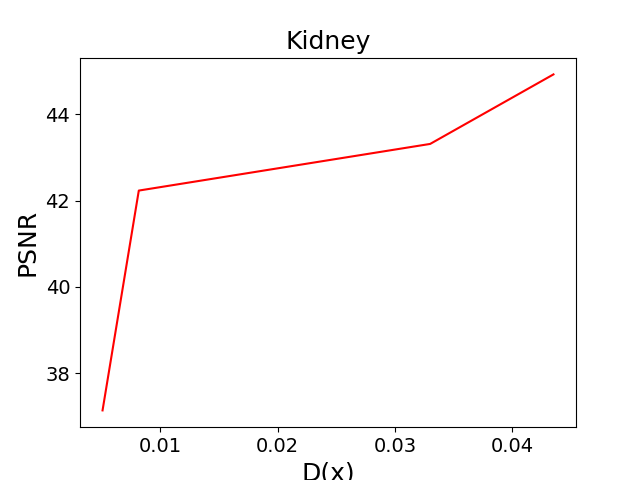}
}
\subfloat{
    \includegraphics[width=0.24\textwidth]{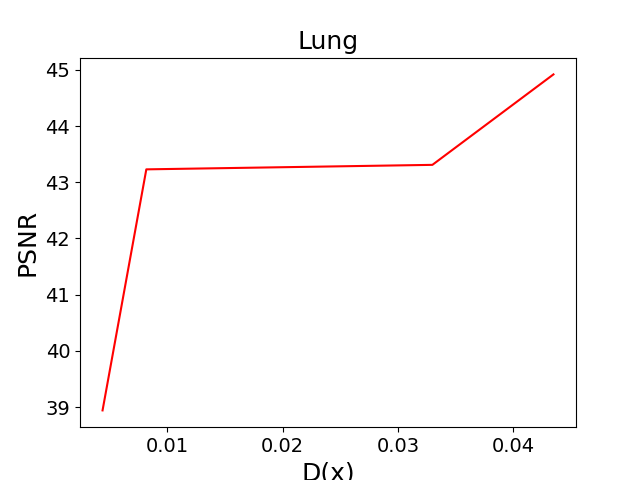}
}
\caption{Relationship between INR's fidelity and the degree of spectrum concentration. $D(x)$ denotes the degree of data's spectrum concentration, the higher means the more concentrated the spectrum. One can easily find a very intuitive positive correlation.}
\label{spectrum concentration}
\end{figure}

As a continuous, differentiable, and parameterized expression, INR has demonstrated its many advantages over traditional discrete expressions. However, INR intrinsically has limited spectrum coverage. For signals with complex frequency components and wide spectral distribution, the reconstruction fidelity of INR will decrease rapidly. In order to quantify this phenomenon, we used data complexity $D(x)$ defined by \cite{yang2022sci}. $D(x)$ measures the proportion of the Top-M frequency components to the total sum of frequencies.

We conducted identical experiments across four distinct datasets. Taking the Brain dataset as an example, we manipulated the data to exhibit varying degrees of spectral concentration by cropping the data from different regions. As depicted in Fig. \ref{spectrum concentration}, there is a noticeable trend: as the spectral concentration of the data intensifies, the difficulty of representation decreases. Concurrently, the reconstruction quality of the Implicit Neural Representation (INR) method shows marked improvement. This observation underscores a clear positive correlation between the degree of spectral concentration and the quality of reconstruction.

\end{document}